\documentclass[letterpaper, 10 pt, conference]{ieeeconf}  

\IEEEoverridecommandlockouts                              
\usepackage{graphics} 
\usepackage{cite}
\usepackage{amsmath,amssymb,amsfonts}
\usepackage{algorithmic}
\usepackage{graphicx}
\usepackage{textcomp}
\usepackage{xcolor}
\usepackage{float}
\usepackage{multirow}
\usepackage{url}

\title{\LARGE \bf
HORUS: A Mixed Reality Interface for Managing Teams of Mobile Robots
}

\author{Omotoye Shamsudeen Adekoya*, Antonio Sgorbissa, Carmine Tommaso Recchiuto
\thanks{* Corresponding author: omotoye.adekoya@edu.unige.it{\tt\small }} 
\thanks{All authors are with DIBRIS Department, RICE Laboratory, University of Genoa, Italy
}
}

\begin{document}

\maketitle
\thispagestyle{empty}
\pagestyle{empty}

\begin{abstract}

Mixed Reality (MR) interfaces have been extensively explored for controlling mobile robots, but there is limited research on their application to managing teams of robots. This paper presents HORUS: Holistic Operational Reality for Unified Systems, a Mixed Reality interface offering a comprehensive set of tools for managing multiple mobile robots simultaneously. HORUS enables operators to monitor individual robot statuses, visualize sensor data projected in real time, and assign tasks to single robots, subsets of the team, or the entire group, all from a Mini-Map (Ground Station). The interface also provides different teleoperation modes: a mini-map mode that allows teleoperation while observing the robot model and its transform on the mini-map, and a semi-immersive mode that offers a flat, screen-like view in either single or stereo view (3D). We conducted a user study in which participants used HORUS to manage a team of mobile robots tasked with finding clues in an environment, simulating search and rescue tasks. This study compared HORUS’s full-team management capabilities with individual robot teleoperation. The experiments validated the versatility and effectiveness of HORUS in multi-robot coordination, demonstrating its potential to advance human-robot collaboration in dynamic, team-based environments.

\end{abstract}

\section{INTRODUCTION}

Multi-robot systems are increasingly deployed in critical applications such as search and rescue \cite{tranzatto_2022_cerberus}, industrial inspection, and exploration, where rapid decision-making and robust situational awareness can be life-saving. However, managing a team of robots in dynamic, complex environments remains challenging due to the limitations of traditional control interfaces. Conventional 2D displays require operators to mentally fuse disparate information—from robot cameras, sensors, and maps—leading to increased cognitive load and potential errors. Mixed Reality (MR) interfaces, which blend real and virtual elements in real time, are emerging as powerful tools in human-robot interaction (HRI). MR extends Augmented Reality (AR) by anchoring virtual content to the real world and enabling two-way interaction between physical and digital objects. In robotics, MR interfaces allow human operators to visualize robots and their data overlaid on the real environment, offering intuitive means to monitor and control them. This capability is especially important for managing teams of mobile robots, where coordinating multiple agents demands rapid assimilation of varied data sources and clear visualization of the mission status \cite{roldn_2017_multirobot}. Overlaying mission data, such as robot locations, paths, obstacles, and sensor feeds, onto real-world views has been shown to improve operator situational awareness, enabling users to detect mission-critical elements more accurately \cite{roldn_2017_multirobot}.

\begin{figure}[h]
\centering
\includegraphics[width=1\linewidth]{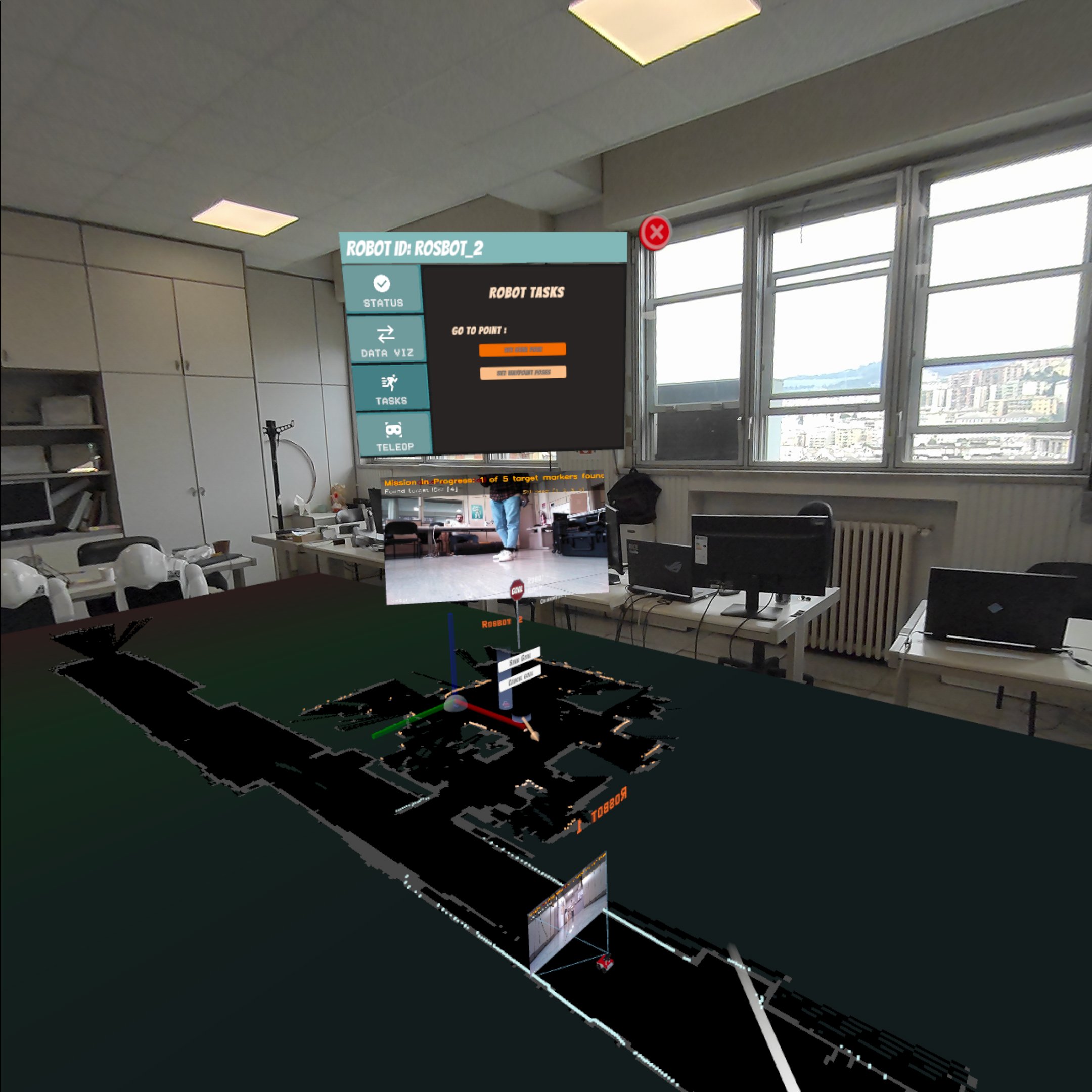}
\caption{The minimap displays two robots: Robot One, which has a projected camera view, and Robot Two, which has an overhead camera view. The Robot Manager Panel for Robot Two is activated, allowing the operator to check the robot’s status, toggle sensor data visualization on or off, assign a task, and initiate teleoperation. Laser scan data visualization is enabled for both robots.}
\label{fig:mini_map}
\end{figure}

Building on these insights and the demonstrated benefits of MR interfaces for single-robot data visualization and control, this paper introduces HORUS (Holistic Operational Reality for Unified Systems), a novel MR interface designed specifically for managing teams of mobile robots. In our implementation, the Husarion ROSBot 2.0 platform serves as the testbed for HORUS. The system leverages a collaborative mini-map that integrates real-time data from each robot to form a unified operational picture. Through this interface, operators can monitor individual robot statuses, visualize sensor data (such as camera feeds, laser-scan outputs, navigation paths, and tf frames), and execute a variety of task commands, including setting goal points, defining waypoints, labeling poses, and drawing navigation plans.

HORUS is organized around a Robot Manager panel that presents dedicated tabs for Robot Status, Robot Data Visualization, Tasks Assignment, and Single Robot Teleoperation for every single robot. Operators interact with each robot’s holographic representation on the mini-map, enabling them to inspect critical parameters and initiate control actions. In scenarios where an operator may need to teleoperate one robot in the team (e.g., manually drive one robot through a tricky area), two distinct teleoperation modes are supported: (i) mini-map teleoperation, where commands are issued by interacting with the robot’s model directly on the map, and (ii) semi-immersive teleoperation, which offers a detailed, flat, screen-like view (in both single and stereo 3D) for precise control while the mini-map remains in the operator’s view. By fusing data from all robots into a common, spatially registered map, HORUS not only enhances situational awareness but also allows fluid switching between high-level task assignment and direct robot control.

This work addresses a significant gap in the literature, where most MR interfaces have predominantly focused on single-robot scenarios or have been confined to simulated environments. In contrast, HORUS has been validated on a real team of wheeled mobile robots, demonstrating its effectiveness in handling the complexities of multi-robot coordination in an operational setting.

\section{RELATED WORKS}
Mixed Reality (MR) interfaces have gained momentum in robotics research due to their potential to overlay rich, spatially registered information onto the physical world. Although numerous single-robot systems have demonstrated the benefits of MR-based teleoperation, data visualization, and task management, extending these capabilities to multi-robot teams introduces greater complexity and new opportunities. This section reviews key developments, ranging from single-robot control to advanced multi-robot coordination, and discusses the existing challenges that our work seeks to address.

Extensive work has shown that Mixed Reality (MR) can reduce cognitive load and improve operator performance in single-robot scenarios, typically by overlaying sensor data, paths, and status indicators within an immersive display. Systems such as ARviz \cite{khoaconghoang_2022_arviz} and iViz \cite{zea_2021_iviz} highlight how headsets or mobile devices augment real-world views with robot information (e.g., camera feeds or 3D trajectories). This intuitive overlay fosters \emph{egocentric} command inputs, enabling operators to “grab” or redirect holographic representations rather than manipulate abstract menus, thereby promoting more natural interactions \cite{walker_2019_robot, allenspach_2023_mixed}.

Building on single-robot successes, researchers have explored drag-and-drop or gesture-based MR interfaces for high-level coordination of multiple robots. Chen \emph{et al.} \cite{chen_2024_a} demonstrated an immersive 3D map where each robot’s avatar is co-localized via multi-agent SLAM; tasks are assigned simply by moving holographic agents to the desired target. Similarly, Kennel-Maushart \emph{et al.} \cite{floriankennelmaushart_2023_interacting} applied a board-game metaphor for multi-robot planning on a HoloLens 2, enabling non-experts to perform task allocation in robot teams with gesture-based interactions and minimal training.

For teams operating in complex or partially occluded spaces, MR interfaces offer rich situational awareness by merging sensor feeds into a single, spatially aligned display. Users can effectively “see through walls” \cite{chen_2024_a}, monitoring a swarm’s positions, planned paths, and camera streams in real time. This unified view supersedes traditional interfaces that require shifting between multiple 2D panels, reducing oversight and enabling timely interventions \cite{roldn_2017_multirobot}.

Finally, some tasks require finer control of specific robots. Immersive systems have shown how an operator can pivot from supervisory modes to direct teleoperation for navigation or manipulation tasks \cite{allenspach_2023_mixed, floriankennelmaushart_2023_interacting}. By embedding holographic cues (e.g., collision warnings or desired end-effector locations), MR can streamline complex telemanipulation for one robot while still tracking the rest of the team’s mission progress.

Despite notable gains, MR systems remain limited by headset field of view, potential tracking drift in large spaces, and the risk of information overload. Furthermore, many solutions rely on small user studies or small robot teams, restricting real-world scalability \cite{roldn_2017_multirobot}.

\vspace{4pt}  Our approach aims to unify these advances into a \emph{single} MR solution that combines intuitive task allocation, single-robot teleoperation, and real-time monitoring for robot teams. By adopting the Meta Quest 3—whose field of view is roughly double that of the Microsoft HoloLens 2—we mitigate visibility constraints and offer more robust inside-out tracking. We also incorporate a flexible “mini-map” interface that does not rely on local environment anchors, allowing operators to manage remote robots (e.g., over a VPN) at large distances. This mini-map, paired with the Quest 3’s handheld controllers, simplifies direct interaction, lowers training time, and supports team scalability: users can quickly view, select, and command multiple agents within a compact 3D layout. Together, these features address common MR challenges while consolidating disparate benefits from prior systems into a more complete solution for multi-robot management.

\section{METHODOLOGY}

This section details the primary components of our architecture. We begin by describing the MR device and software stack, highlighting the advantages of the Meta Quest 3 headset. Next, we outline our robotics platform and explain how each unit contributes to multi-robot SLAM and navigation through individual mapping and a custom map-merging approach. Finally, we present the Robot Manager panel—a comprehensive interface that organizes status monitoring, sensor data visualization, mission planning, and teleoperation into dedicated, user-friendly tabs.

\begin{figure*}[h]
\centering
\includegraphics[width=0.62\linewidth]{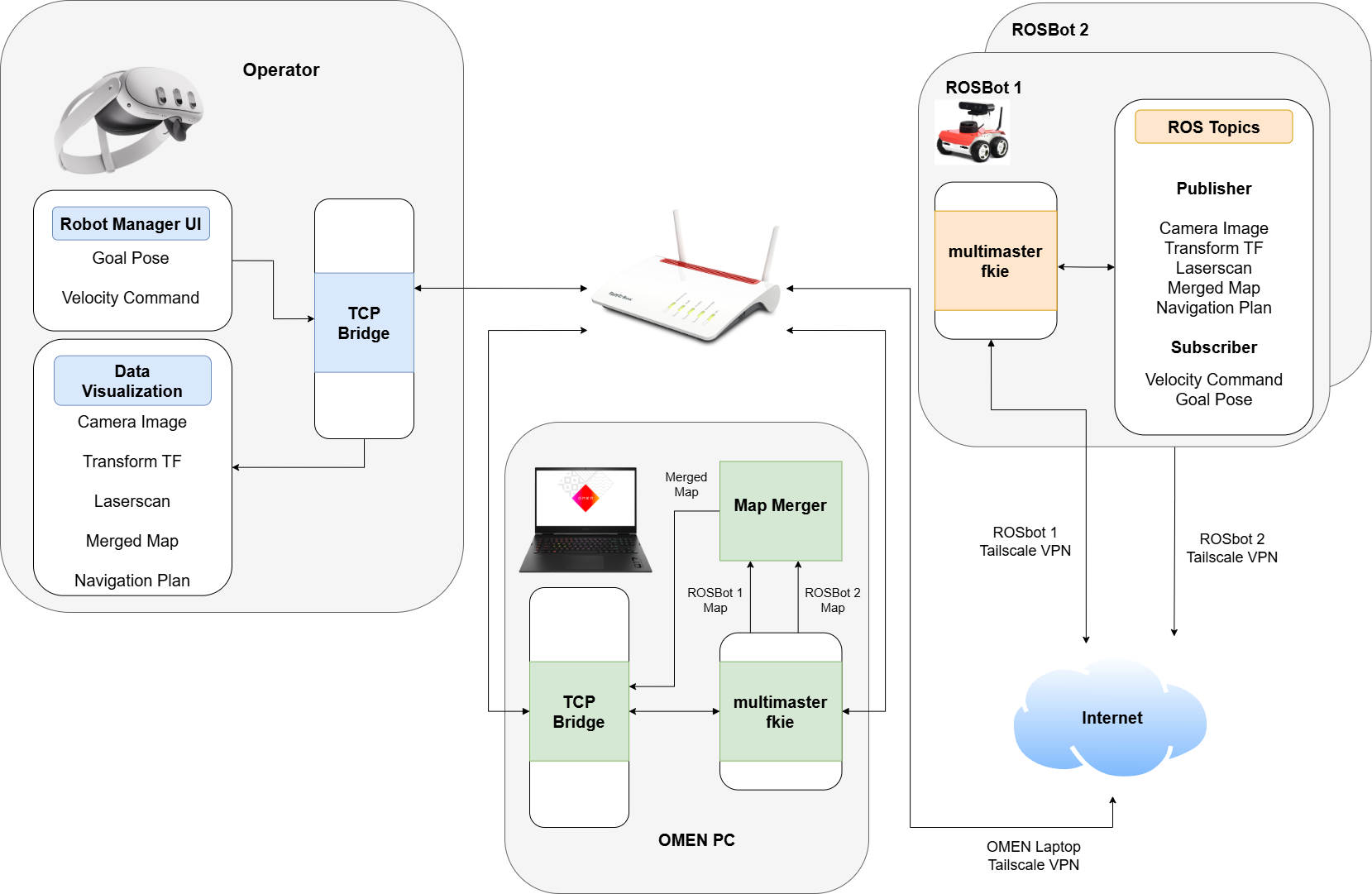}
\caption{System Architecture Diagram}
\label{fig:system_arch}
\end{figure*}

\subsection{MR Device and Software Stack}
We employ the Meta Quest 3 headset for its color passthrough technology and substantially wider field of view compared to devices such as the HoloLens 2. This expanded FOV offers more immersive mixed-reality visuals, allowing users to see larger portions of the real environment along with the augmented content. We developed our interface in Unity 2022.3.34f1, integrating the Meta XR All-in-One SDK v72 for creating interaction elements in the interface and sensor visualization prefabs. The entire application is compiled directly onto the Quest 3 for untethered operation, eliminating the need for wired connections to a workstation.

A Unity TCP Connector package was used to communicate with the robots through ROS topics. This package was developed by Unity to enable a connection with a ROS master through TCP rosbridge, as long as both the headset and the ROS machine are connected to the same local network. The HORUS Bridge node runs on a remote machine, synchronizing Unity-based user commands with each robot’s onboard ROS modules. It is an extension of the Unity TCP Endpoint ROS package and handles the bridge between machines and specific messages from the interface on the headset. Since each robot runs a separate ROS Master, the \emph{multimaster\_fkie} package\footnote{\url{https://github.com/fkie/multimaster_fkie}} enables multi-master operation. In contrast to a single-master configuration that routes all robot data through one entity (risking congestion and a single point of failure), the multi-master approach replicates only necessary topics among robots, which scales better for increased team sizes.

For fully remote deployments, we use Tailscale VPN to seamlessly connect operators and robots over the internet. Tailscale’s lightweight, peer-to-peer mesh structure makes it straightforward to establish secure links without custom firewall or port configurations. Tailscale creates a secure mesh network using WireGuard that directly connects devices anywhere with zero configuration, works through firewalls without dedicated servers, authenticates through existing providers, and only encrypts device-to-device traffic. This setup makes devices function as though they are on the same local network with minimal effort.

\subsection{Robotics Platform}
We validated our system on the Husarion ROSbot 2.0, a compact, differential-drive robot equipped with a forward-facing LiDAR, wheel encoders, and an onboard ARM7L-based controller (Asus Tinker Board) for ROS 1. Although it lacks advanced 3D sensing, the ROSbot’s 2D LiDAR and odometry are sufficient for basic navigation and mapping. Its moderate footprint and maneuverability also make it suitable for multi-robot experiments in typical indoor environments.

\subsection{Multi-Robot SLAM and Navigation}

Multi-robot SLAM research increasingly highlights full 3D approaches with loop-closure detection as state-of-the-art solutions for building globally consistent maps. Such systems, however, often rely on more capable hardware (e.g., depth cameras or high-end LiDAR) and stronger computing resources to handle complex feature extraction and large-scale pose graph optimization. Given the limited sensing and onboard processing of the ROSbot 2.0, our implementation favors a streamlined approach that still enables multi-robot collaboration without overburdening the hardware.

Specifically, each ROSbot independently runs GMapping\footnote{\url{https://wiki.ros.org/gmapping}} to create a local occupancy grid map from its 2D LiDAR and odometry. We unify these local maps through a custom script that applies a coarse alignment (obtained via ROS TF) and then refines it with a phase-correlation step using OpenCV. The resulting merged occupancy grid leverages overlapping observations, mitigating individual odometry drift and improving overall map detail.

For navigation, each robot shares this consolidated map within its navigation stack, using the Timed Elastic Band (TEB) local planner \cite{rosmann2017integrated} for real-time collision avoidance and dynamic path adaptation. By combining independent local mapping, robust map merging, and a reactive local planner, we achieve a cohesive multi-robot SLAM and navigation framework that remains within our target platform’s computational budget, effectively improving operational reliability and accuracy for teams of low-cost mobile robots.

\subsection{Robot Manager Panel}
\label{sec:robot_manager}

 \begin{figure}[h]
\centering
\includegraphics[width=0.8\linewidth]{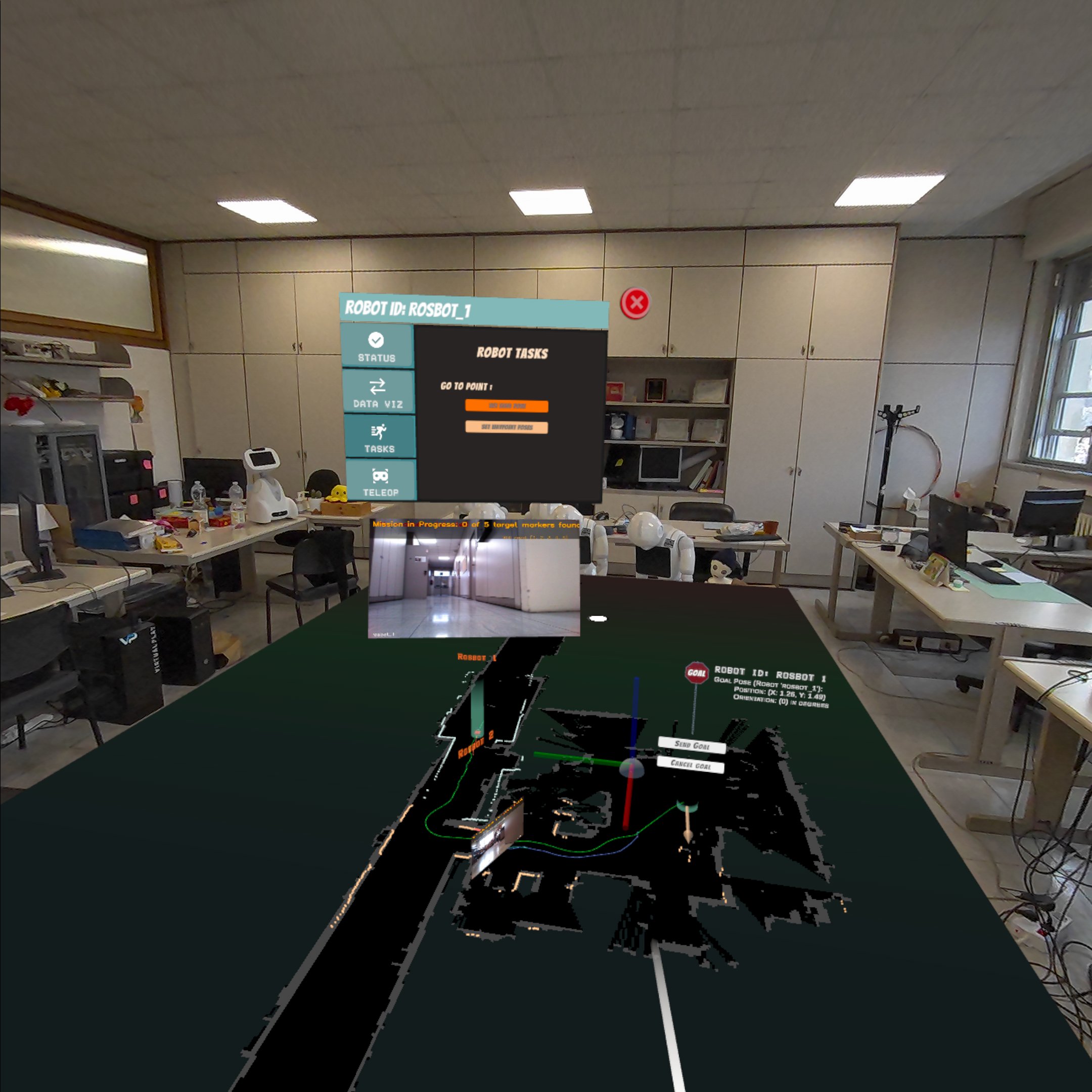}
\caption{The minimap with all sensor data visualization switched on for both robots in a typical use-case scenario.}
\label{fig:path_viz}
\end{figure}

A core component of our interface is the \emph{Robot Manager} panel, which appears whenever a user selects a robot on the mini-map. A dedicated \emph{PanelPositioner} script keeps this panel facing the operator’s headset, allowing users to walk around the mini-map without losing access to the robot’s controls. The panel is divided into four primary tabs, each targeting a different aspect of multi-robot management.

\textbf{1) Status.}
This tab displays essential robot properties such as battery level, velocity, and current pose. It also shows \emph{task completion status}, indicating whether the robot is actively navigating to a goal or waiting for new commands. Providing real-time feedback on these parameters allows the operator to quickly identify and address potential issues (e.g., low battery or stalled navigation).

\textbf{2) Data Viz.}
In this section, the operator can enable or disable various sensor overlays, including LiDAR scans and camera feeds. For LiDAR, the interface renders colored scan points around the robot; points turn \emph{red} if they are within 0.5 m of an obstacle, warning operators of potentially unsafe proximity.

\emph{Camera Modes.} Two modes are available for the robot’s live video stream:
\begin{itemize}
  \item \textbf{Projected} — The feed is pinned to the robot’s physical camera frame, as if the operator is “looking through” the robot’s onboard camera.
  \item \textbf{Overhead} — A larger, floating window appears just above the robot on the mini-map. It also uses the \emph{PanelPositioner} script to maintain front-facing alignment with the user. This mode is especially useful for mini-map teleoperation, enabling a clearer view of the environment without occluding the mini-map itself.
\end{itemize}

\textbf{3) Tasks.}
This tab manages mission-level commands that the robot should execute:
\begin{itemize}
  \item \emph{Set Goal Pose.}
  When enabled, a goal-pose prefab appears on the map where the user taps or clicks. The prefab includes an orientation arrow that can be dragged left or right to set the heading, with numerical orientation updates. This workflow lets the operator define precise navigation objectives using minimal input.

  \item \emph{Set Waypoint Poses.}
  Operators can designate multiple waypoints on the mini-map, creating a more controlled path. This feature allows for incremental route planning and smooth navigation by specifying intermediate positions along a desired route.

  \item \emph{Label Robot Pose.}
  Users can annotate specific locations in the environment (e.g., “Wounded Victim Here”). In search-and-rescue scenarios, such labeling helps coordinate responses by highlighting key areas for the rest of the team.

  \item \emph{Draw Navigation Plan.}
  Operators can sketch a custom global path on the mini-map, and the robot’s local planner will follow the path as closely as possible, ultimately converging on the end pose. This is useful when the operator wants the robot to pass through specific areas rather than computing the shortest path.
\end{itemize}

\textbf{4) Teleoperation}

 \begin{figure}[h]
\centering
\includegraphics[width=0.75\linewidth]{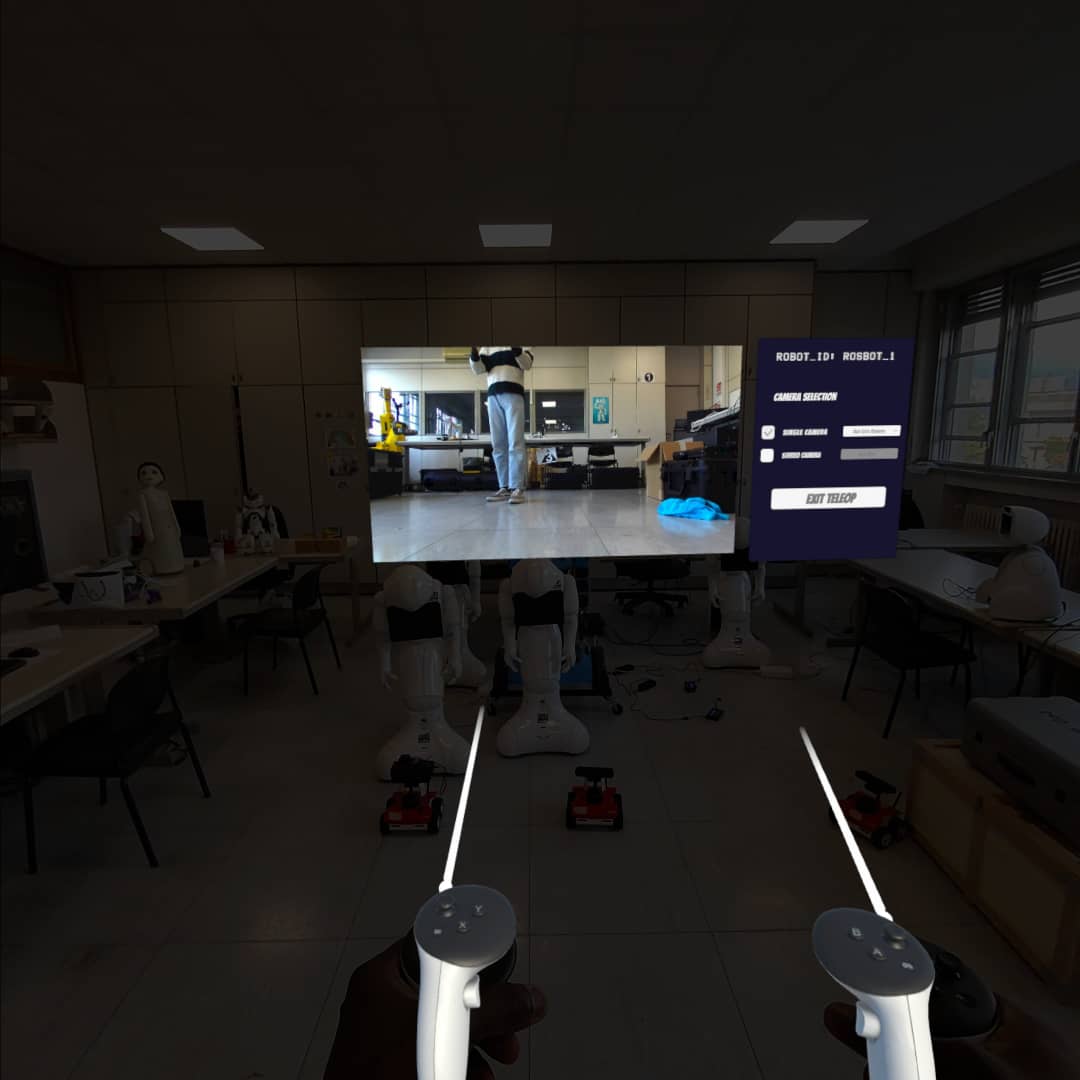}
\caption{Semi-immersive teleoperation mode.}
\label{fig:teleop}
\end{figure}

Two teleoperation modes are provided, each using the Meta Quest 3 controllers for linear and angular velocity inputs:
\begin{itemize}
  \item \emph{Mini-map Teleoperation.}
  The robot is driven from a third-person perspective on the mini-map, similar to controlling an RC car in a top-down game. The left joystick controls linear speed, and the right joystick handles angular speed. Pressing the B button increments angular speed by 0.1, and pressing again decrements by 0.1. Similarly, Y increases linear speed, and X decreases it. Holding the left grip activates linear motion; holding the right grip enables rotation. Pressing both grips resets the speeds to their defaults. While camera visuals are optional here, the overhead camera view is recommended for improved environmental awareness.

  \item \emph{Semi-Immersive Teleoperation.}
  This mode provides a larger, 2D or stereoscopic camera feed from the robot in front of the user. The same joystick and button bindings apply, but the view is more immersive, ideal for maneuvering through tight corridors or cluttered areas. In the case of stereo-capable cameras, each eye receives the corresponding viewpoint to achieve a 3D effect and enhance depth perception.
\end{itemize}

By organizing robot functions into four easy-to-use tabs, operators can smoothly switch between checking statuses, adjusting sensors, planning missions, and directly controlling robots—all within the same MR interface. This setup not only streamlines workflows but also makes it easier for users to manage multiple robots at once, with less time spent learning how to use the system.

\section{USER STUDY}

To assess the usability and performance of our multi-robot Mixed Reality (MR) interface, we conducted a user study in which participants explored a corridor of offices and labs to locate five hidden ArUco tags \cite{garrido2014automatic}. Two groups of 10 participants each used different system configurations: (1) the full HORUS application (\emph{HORUS Group}), and (2) a standalone, semi-immersive teleoperation interface (\emph{Teleop-Only Group}).

\subsection{Study Setup}
All trials took place on a lab floor that opens into a corridor with multiple adjacent offices and labs. Five ArUco tags were placed in various rooms off the corridor. Each participant controlled two ROSbot 2.0 robots connected to a Meta Quest 3 headset (via an HP Omen laptop running the HORUS Bridge and a Tailscale VPN).

\textbf{1) HORUS Group.}
Participants in this group had access to the full HORUS application, excluding the semi-immersive teleoperation feature. They could assign navigation tasks by setting incremental goal poses along the corridor to build a shared SLAM-based map. Upon reaching doorways or areas requiring fine control, operators switched to the mini-map teleop mode, optionally enabling an overhead camera feed for better visibility. The ArUco identification node overlays detection status (“X of 5 items found for the MISSION”) onto the live camera feed; when a tag is detected by either robot, the counter updates and appears in the operator’s headset.

\textbf{2) Teleop.-Only Group.}
Participants in the second group used a dedicated “semi-immersive” application with a large camera feed and direct velocity control via joysticks and button-based speed adjustments, but no autonomous goal-setting or mini-map interface. Operators could toggle between the two robots’ video streams. As in the HORUS group, the detection counter was synchronized between both robots and updated whenever a tag was found.

\textbf{Procedure.}
Each participant received a short training session in an open space to learn basic motion control and, where applicable, goal-based navigation. Both robots started inside the lab, and participants navigated them through the corridor and adjoining rooms in search of the five ArUco tags. Once all tags were located (“5 of 5 items found”), participants returned the robots to the initial lab position.

\subsection{Measurements}
We focused on four primary metrics to evaluate the two interface conditions:

\begin{itemize}
  \item \textbf{Task Completion Time:} From when the user initiated corridor navigation until all tags were detected and both robots returned to the lab.
  \item \textbf{System Usability Scale (SUS):} A post-task questionnaire assessing overall interface usability.
  \item \textbf{NASA TLX:} A measure of perceived workload across factors like mental demand, physical demand, and effort.
  \item \textbf{Simulator Sickness Questionnaire (SSQ):} Administered before and after the study to capture any VR/AR-induced discomfort.
\end{itemize}

\subsection{Study Results}

Our user study revealed notable differences between the HORUS and Teleop-Only interfaces across multiple metrics. Overall, the HORUS system demonstrated superior performance and usability compared to the standalone teleoperation interface. In the following discussion, given the amount of data, the results were compared using a parametric test for statistical analysis (i.e., Mann-Whitney U) \cite{mcknight2010mann}.

\subsubsection{Training Requirements}

As expected, the HORUS interface required more thorough training than the Teleop-Only interface. Teleop-Only participants completed training in about 7 minutes, focusing mainly on basic robot control. HORUS participants had an 18-minute training session covering both goal-based navigation and mini-map teleop. Despite this additional time, participants rated HORUS as significantly more usable in post-task assessments, suggesting its intuitive design facilitates effective use of advanced features.

\subsubsection{Task Completion Time}

Analysis showed that HORUS users completed the ArUco tag detection mission significantly faster than Teleop-Only users. Table \ref{tab:completion_time} summarizes the findings.

\begin{table}[h]
\centering
\caption{Task Completion Time Comparison}
\label{tab:completion_time}
\begin{tabular}{lccc}
\hline
\textbf{Interface} & \textbf{Mean (min:sec)} & \textbf{Median (min:sec)} & \textbf{Std. Dev. (sec)} \\
\hline
HORUS & 8:42 & 8:28 & 62.3 \\
Teleop-Only & 11:17 & 10:49 & 87.5 \\
\hline
\multicolumn{4}{l}{$t(18) = 4.32, p < 0.001$, Cohen's $d = 1.93$} \\
\end{tabular}
\end{table}

This time advantage ($p < 0.001$, large effect size $d = 1.93$) demonstrates HORUS’s effectiveness in enhancing mission efficiency. Goal-based navigation and map-building let operators systematically explore and coordinate more effectively than direct teleoperation alone, even with only 20 minutes of training.

\subsubsection{System Usability Scale (SUS)}

\begin{figure}[h]
\centering
\includegraphics[width=1\linewidth]{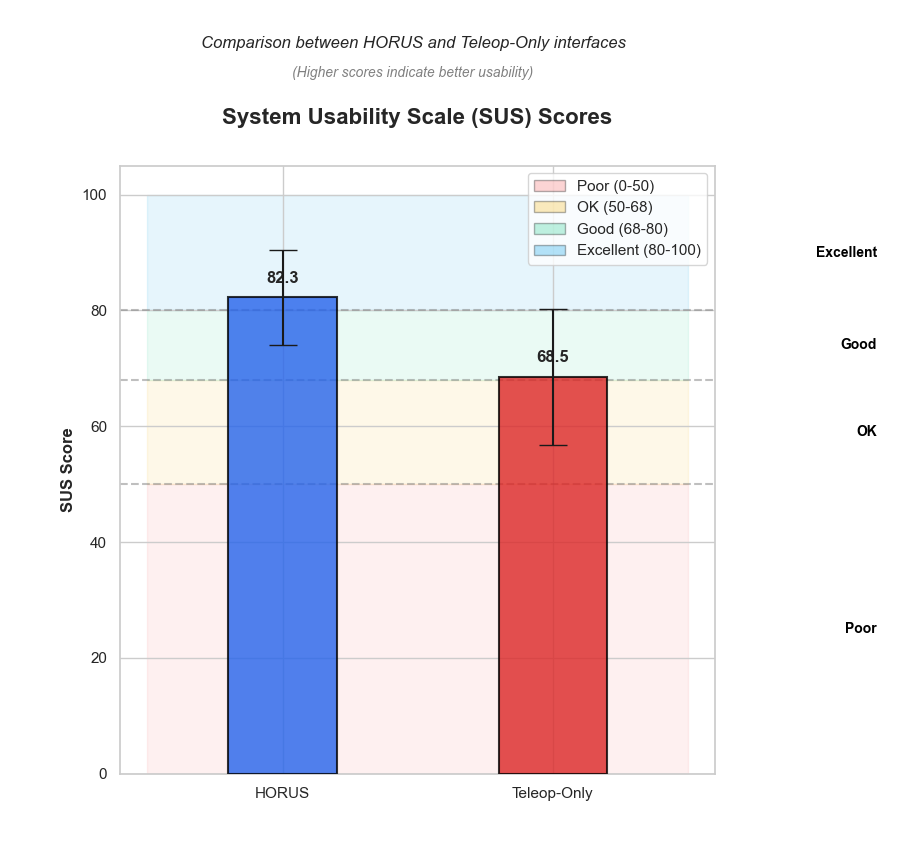}
\caption{System Usability Scale scores for HORUS and Teleop-Only interfaces. Error bars represent standard error.}
\label{fig:sus_scores}
\end{figure}

SUS scores indicated a clear usability advantage for HORUS over Teleop-Only. Figure \ref{fig:sus_scores} shows the comparison.

HORUS achieved a mean SUS score of 82.3 (SD = 8.2), categorized as “excellent” usability, while Teleop-Only scored 68.5 (SD = 11.7), rated “good.” The difference was statistically significant ($t(18) = 3.12, p = 0.006$), indicating that participants found HORUS more usable. Table \ref{tab:sus_items} details responses to individual SUS items.

\begin{table}[h]
\centering
\caption{Mean Scores for Individual SUS Items (Scale: 1-5)}
\label{tab:sus_items}
\begin{tabular}{lcc}
\hline
\textbf{Item} & \textbf{HORUS} & \textbf{Teleop-Only} \\
\hline
I would like to use this system frequently & 4.2 & 3.3 \\
The system was unnecessarily complex & 1.7 & 2.9 \\
The system was easy to use & 4.1 & 3.2 \\
I would need technical support to use & 1.8 & 2.7 \\
Functions were well integrated & 4.3 & 3.1 \\
Too much inconsistency & 1.6 & 2.8 \\
Most would learn to use quickly & 4.3 & 3.4 \\
Found the system cumbersome & 1.5 & 2.6 \\
Felt confident using the system & 4.2 & 3.0 \\
Needed to learn a lot before using & 1.9 & 2.5 \\
\hline
\end{tabular}
\end{table}

Participants especially appreciated HORUS’s integrated functions and reported higher confidence than with Teleop-Only. They also noted that new users would learn HORUS quickly despite its advanced features.

\subsubsection{NASA TLX Workload Assessment}

\begin{figure}[h]
\centering
\includegraphics[width=1\linewidth]{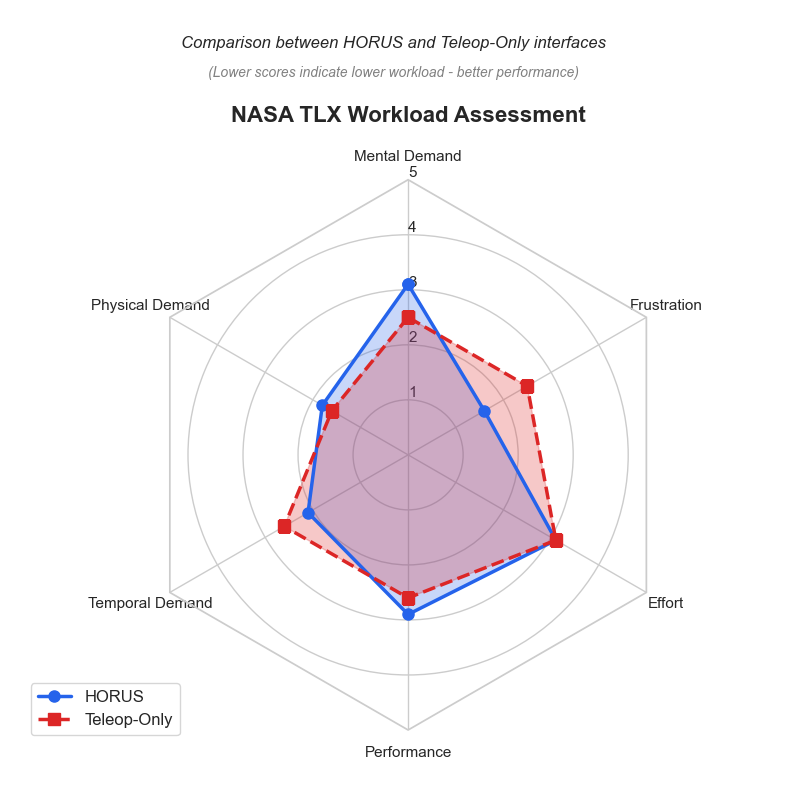}
\caption{NASA TLX workload assessment comparison (lower scores indicate lower workload).}
\label{fig:nasa_tlx}
\end{figure}

The NASA TLX revealed important differences in perceived workload. Figure \ref{fig:nasa_tlx} compares the workload dimensions. While the overall workload was similar between HORUS (M = 2.43, SD = 0.76) and Teleop-Only (M = 2.48, SD = 0.84), the distribution of scores across dimensions highlighted key advantages for HORUS. Notably, frustration was significantly lower with HORUS ($t(18) = 2.21, p = 0.04$) (Table \ref{tab:nasa_tlx}).

\begin{table}[h]
\centering
\caption{NASA TLX Workload Dimensions (Scale: 1-5, lower is better)}
\label{tab:nasa_tlx}
\begin{tabular}{lcccc}
\hline
\multirow{2}{*}{\textbf{Dimension}} & \multicolumn{2}{c}{\textbf{HORUS}} & \multicolumn{2}{c}{\textbf{Teleop-Only}} \\
\cline{2-5}
 & \textbf{Mean} & \textbf{SD} & \textbf{Mean} & \textbf{SD} \\
\hline
Mental Demand & 3.1 & 1.45 & 2.5 & 1.08 \\
Physical Demand & 1.8 & 1.14 & 1.6 & 0.84 \\
Temporal Demand & 2.1 & 0.57 & 2.6 & 1.65 \\
Performance & 2.9 & 2.38 & 2.6 & 1.65 \\
Effort & 3.1 & 1.45 & 3.1 & 1.37 \\
Frustration & 1.6 & 0.70 & 2.5 & 1.43 \\
\hline
Overall Workload & 2.43 & 0.76 & 2.48 & 0.84 \\
\hline
\end{tabular}
\end{table}

Although HORUS introduced slightly higher mental demand, it significantly reduced frustration, contributing to a more positive user experience.

\subsubsection{Simulator Sickness Questionnaire (SSQ)}

\begin{figure}[h]
\centering
\includegraphics[width=1\linewidth]{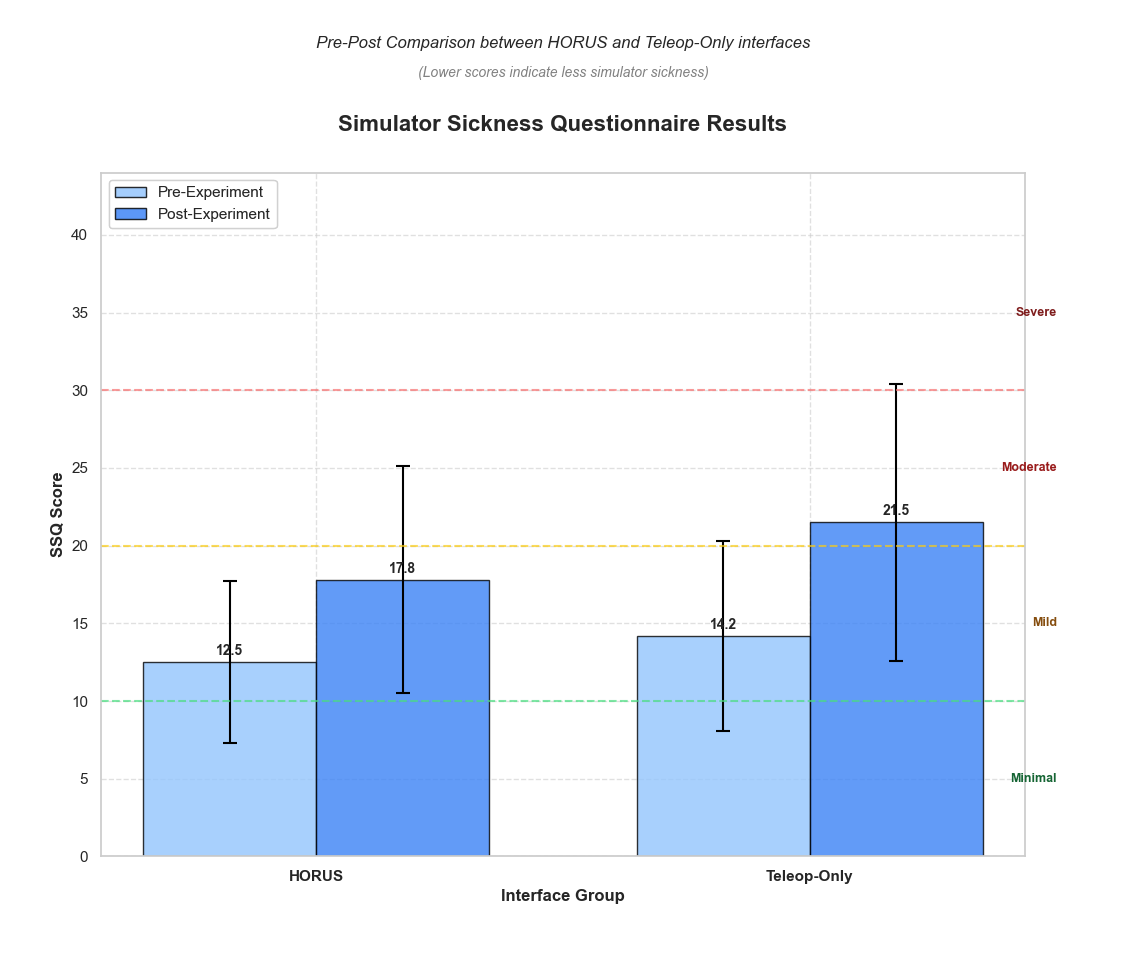}
\caption{Pre-post changes in SSQ scores for both interfaces.}
\label{fig:ssq_scores}
\end{figure}

The SSQ was given both before and after the experiment to gauge discomfort from MR use. Both interfaces resulted in minimal simulator sickness, with no significant group differences. Figure \ref{fig:ssq_scores} illustrates the pre-post SSQ changes.

Neither interface induced clinically significant symptoms, indicating that the MR visualization methods were comfortable for extended usage.

\vspace{4pt}. As qualitative data, a key benefit of the HORUS interface was its support for coordinated multi-robot operations. HORUS users tended to employ complementary search strategies, assigning robots to distinct areas to maximize coverage. In contrast, Teleop-Only participants often focused sequentially on one robot at a time, reducing exploration efficiency.

Overall, the HORUS application demonstrated several advantages over the Teleop-Only teleoperation. It was approximately 23\% faster in task completion, highlighting how its goal-based navigation significantly boosts mission speed. Participants also found HORUS to be more intuitive, as evidenced by substantially better SUS scores, despite its more advanced features. The frustration levels were notably lower in the HORUS group, as reflected in the significantly lower NASA TLX frustration scores. Additionally, the shared map features of HORUS fostered improved multi-robot coordination, encouraging parallel tasking and resulting in more effective coverage of the environment.

The mini-map teleop mode was especially beneficial, merging situational awareness with precise manual control. This dual approach—combining autonomous navigation for efficiency with manual teleop for precision—offers a step forward over purely teleoperated interfaces.

\section{CONCLUSIONS AND FUTURE WORKS}

This paper presented HORUS, a Mixed Reality interface for managing teams of mobile robots that integrates visualization, task allocation, and flexible control modalities. Our user study demonstrated substantial advantages over standard teleoperation: faster task completion, superior usability scores, reduced operator frustration, and improved multi-robot coordination. While HORUS required a modestly longer training period (20 minutes vs. 7 minutes), the benefits in performance and user satisfaction justified the additional training \cite{kennelmaushart2023interacting, chen2023mixed}.

Future work will focus on several areas:

\begin{itemize}
    \item \textbf{Heterogeneous Robot Teams:} Incorporating legged robots (e.g., Spot) and aerial vehicles (e.g., Airvolute, DJI drones) to enable multi-domain missions \cite{delmerico2022spatial}.
    \item \textbf{Advanced Control and Collaboration:} Implementing more complex trajectory-planning tools, multi-operator functionality, and AI copilot systems to assist operators \cite{kennelmaushart2023interacting, suzuki2022augmented}.
    \item \textbf{3D Mapping and Enhanced Situational Awareness:} Adding advanced 3D mapping techniques for improved multi-robot SLAM \cite{zhang2023overview, oleynikova2017voxblox} and offering operators richer environmental context for better decision-making \cite{reardon2019communicating, cramariuc2023maplab}.
\end{itemize}

Future testing will span various operational scenarios, including simulated disaster environments, industrial inspections, and collaborative construction tasks. These advancements aim to further enhance human-robot collaboration in complex real-world operations \cite{reardon2018come, chen2023mixed}.

\bibliography{references}

\end{document}